# 6422 | Sinkage Study in Granular Material for Space Exploration Legged Robot Gripper

Arthur Candalot [a,*,†], James Hurrell [a,†], Malik-Manel Hashim [b], Brigid Hickey [c], Mickael Laine[a], Kazuya Yoshida [a]

[a] *Department of Aerospace Engineering, Tohoku University, Sendai, Miyagi, Japan*

[b] *Technische Universität Berlin, 10623 Berlin, Germany*

[c] *University at Buffalo, NY 14068, United States*

** Corresponding author: acandalot@dc.tohoku.ac.jp*

*†These authors contributed equally*

**A B S T R A C T**

Wheeled rovers have been the primary choice for lunar exploration due to their speed and efficiency. However, deeper areas, such as lunar caves and craters, require the mobility of legged robots. To do so, appropriate end effectors must be designed to enable climbing and walking on the granular surface of the Moon.

This paper investigates the behavior of an underactuated soft gripper on deformable granular material when a legged robot is walking in soft soil. A modular test bench and a simulation model were developed to observe the gripper sinkage behavior under load.

The gripper uses tendon-driven fingers to match its target shape and grasp on the target surface using multiple micro-spines. The sinkage of the gripper in silica sand was measured by comparing the axial displacement of the gripper with the nominal load of the robot mass. Multiple experiments were performed to observe the sinkage of the gripper over a range of slope angles. A simulation model accounting for the degrees of compliance of the gripper fingers was created using Altair MotionSolve software and coupled to Altair EDEM to compute the gripper interaction with particles utilizing the discrete element method. After validation of the model, complementary simulations using Lunar gravity and a regolith particle model were performed.

The results show that a satisfactory gripper model with accurate freedom of motion can be created in simulation using the Altair simulation packages and expected sinkage under load in a particle-filled environment can be estimated using this model. By computing the sinkage of the end effector of legged robots, the results can be directly integrated into the motion control algorithm and improve the accuracy of mobility in a granular material environment.

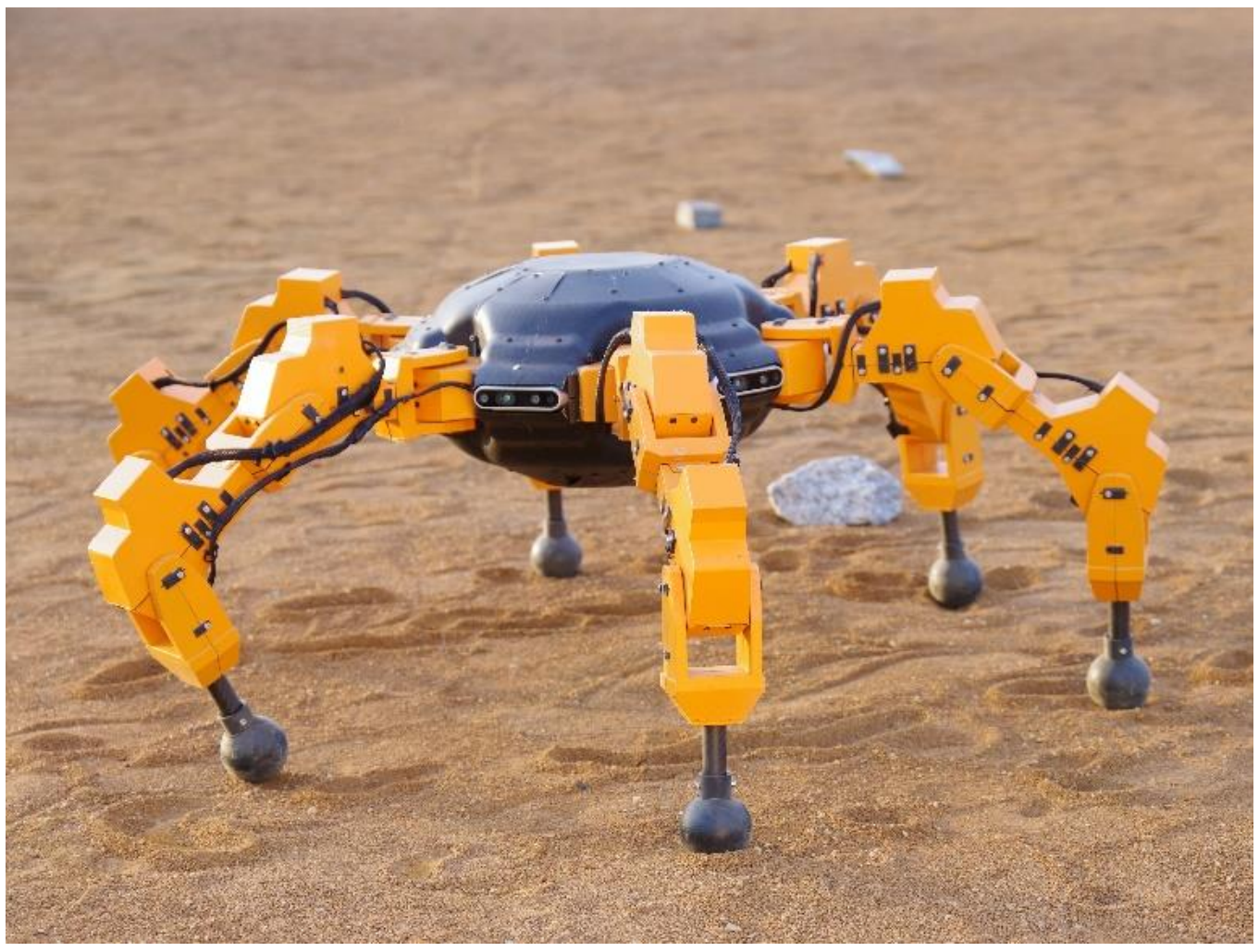

*Fig. 1 SCAR-E hexapod robot walking in dry soil.*

## 1. Introduction

Since the beginning of space exploration, observation of the surface of multiple celestial bodies has been achieved in our solar system. Thanks to pioneer missions, valuable data and samples were collected, and the feasibility of space exploration was confirmed. In recent years, the discovery of high value resources such as water on the Moon (Li and Milliken, 2017) has motivated the development of more advanced robots. These resources will become a key factor in the creation of extraterrestrial infrastructures and potentially support deeper space exploration.

### *1.1. Wheeled and legged space robots*

The first planetary missions focused mainly on surface exploration using wheeled rovers (Rodríguez-Martínez et al., 2024; Welch et al., 2013). Such robots allow for reliable locomotion on flat ground and can overcome minor obstacles. However, the sandy surface of the Moon or Mars creates challenging environments with the risk of causing wheels to slip and sink, especially on steeper terrain. New areas of interest, such as craters, caves, and steeper terrain, cannot be accessed with conventional wheeled rovers. For this reason, alternative approaches have been considered to improve mobility in uneven terrain. Currently, research on snake-like locomotion in granular environment (Marvi et al., 2014), crab-shaped robots (Graf Nicole M. and Behr, 2019), or leg-wheel hybrid robots like RHEX (Saranli et al., 2001) is being conducted for Earth-based applications, although the technology may produce successful results in space.

Legged robots are also becoming a viable solution for the exploration of extreme environments due to their increased ability to climb and overcome obstacles. Recent quadrupeds (Di Carlo et al., 2018) and biped robots (Komizunai et al., 2010) usually employ spherical or flat appendices as feet, which provides stability even on softer terrain (Yang et al., 2024).

Planetary exploration robots (Parness et al., 2017; Uno et al., 2021) are often equipped with spine grippers to grasp the rough surface of rocks, allowing them to descend into lunar caves. Current studies focus on the climbing performances of gripping systems, yet the gripper will be used as a foot during walking. In a granular environment, the foot is expected to sink at each step under the load of the robot, potentially hindering their locomotion. During field testing of the SCAR-E hexapod robot platform (Fig. 1), currently under development at the Space Robotics Laboratory of Tohoku University, it has been observed that if the depth of the sinkage is not taken qualitatively into account in the autonomous GAIT algorithm, discrepancies occur between the computed position and the real position, causing locomotion failure. Therefore, the terramechanic behavior of the equipped end effector must be assessed to ensure safe autonomous navigation in a granular environment.

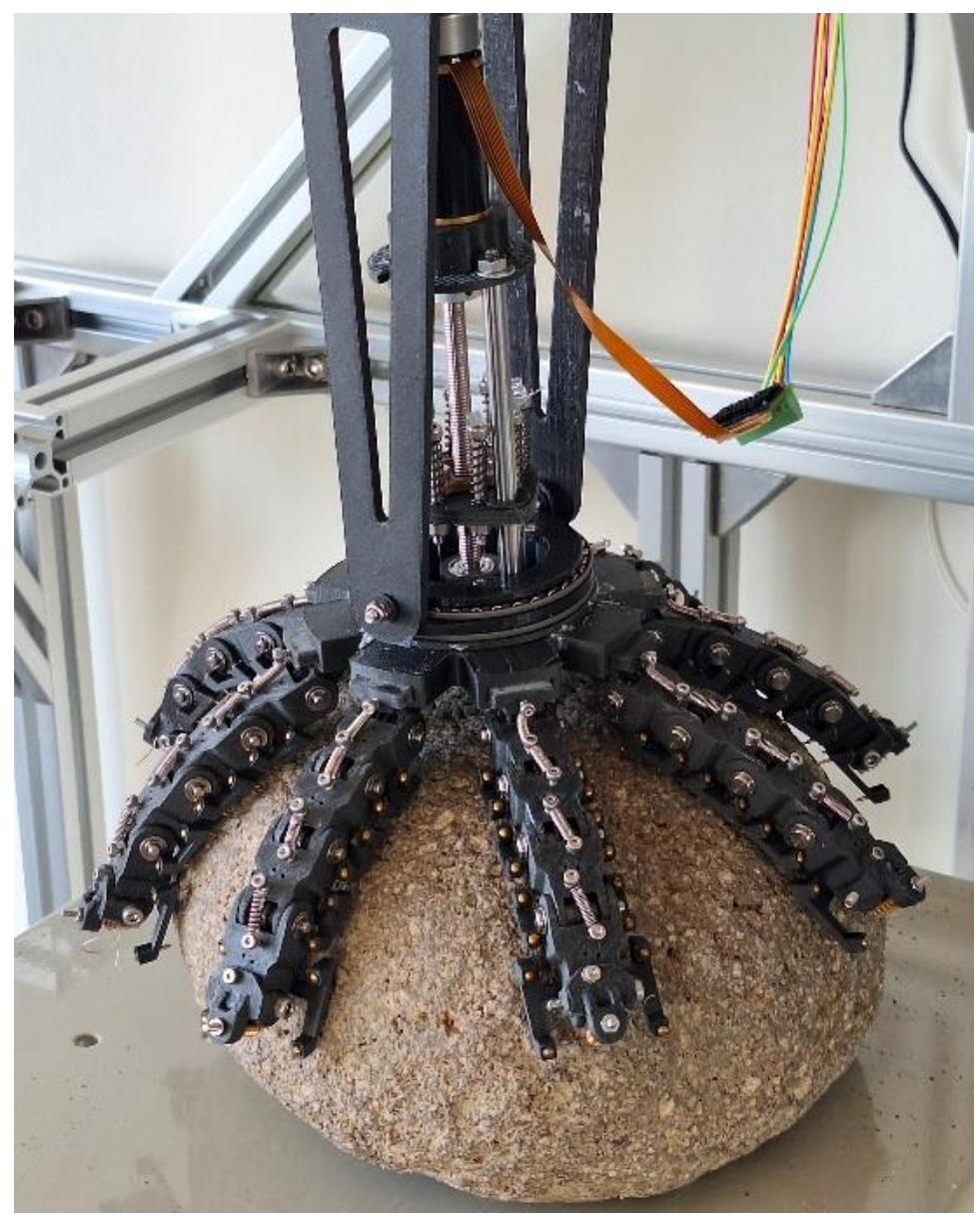

*Fig. 2 Gripper side view*

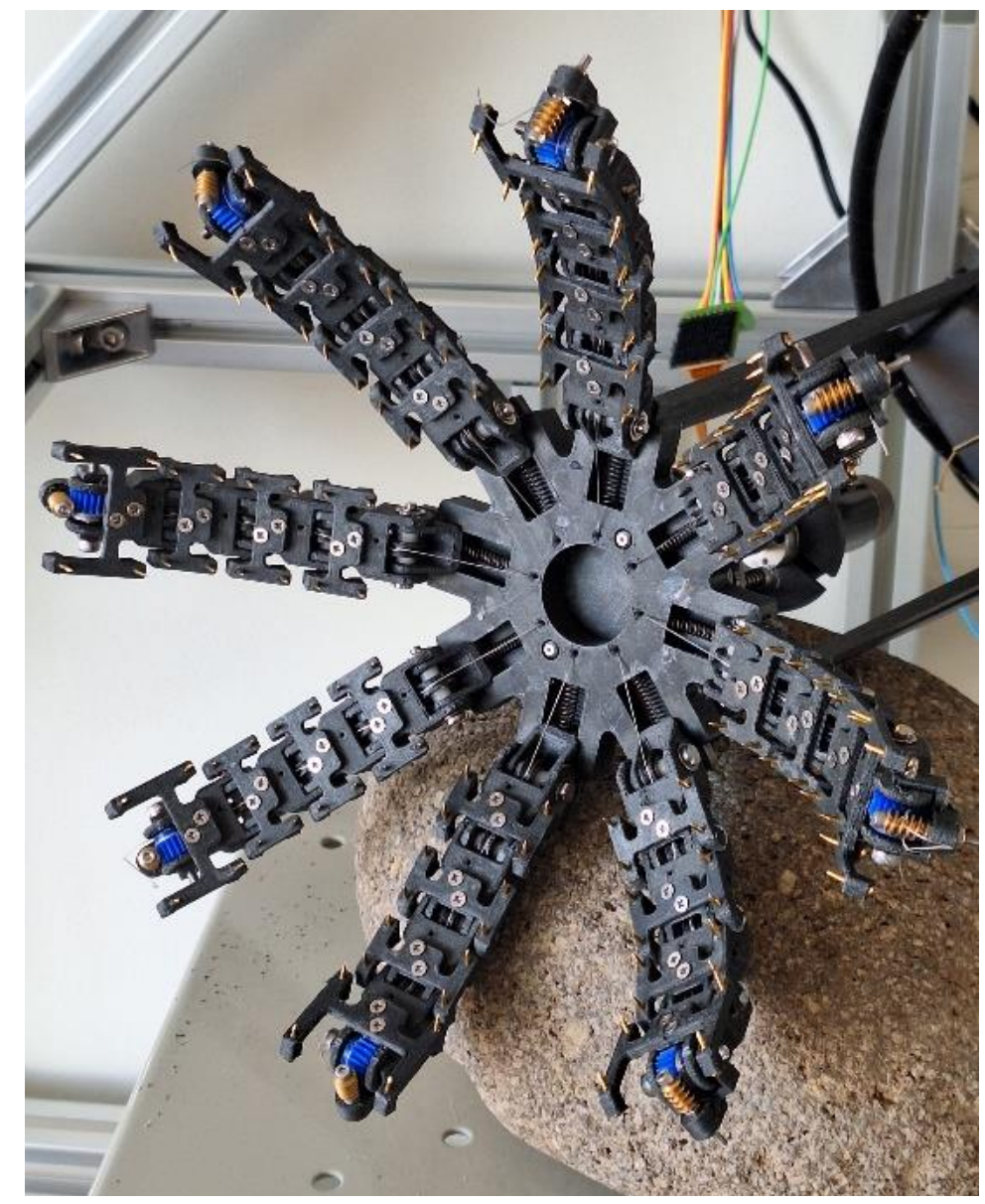

*Fig. 3 Gripper bottom view*

## 1.2. Pressure-sinkage model

The wheel-soil interaction has been an important topic of interest in the development of wheeled vehicles for off-road applications to improve mobility. One of the foci resides on the quantification of expected sinkage during locomotion. Bekker et al. (1962) first introduced a model linking the sinkage to pressure applied and the granular material properties. The model was further refined by Wong & Reece (1967) to include the dynamics of wheel rotation. Grouser wheels (Nagaoka et al., 2020) have been used to improve contact with the soil, but slipping and dynamic sinking still pose a challenge due to shearing of the material, especially on inclined terrain (Ishigami et al., 2006; Oettershagen et al., 2019).

In the case of legged robots, the end effector compresses the granular material and then applies a tangential force as a step is made. The sinkage is therefore induced by the dynamic entry of the end effector into the soil, as well as the normal and tangential pressure applied (Yang et al., 2024). For an equivalent vehicle weight, a stepping motion may generate less shearing than a conventional grouser wheel, allowing a legged robot to overcome steeper slopes.

## 1.3. Gripping system

The proposed gripper (Fig. 2 and Fig. 3) for this study is a soft gripping system with compliant fingers to match the shape of its target and maximize the contact area for improved performances (Candalot et al., 2024). It is constituted of 8 tendon-driven fingers, each divided into 4 rotating phalanges. A tether is attached to the extremity of the finger and is guided through a series of pulleys linked to each phalanx, as shown in Fig. 4. This architecture was previously used in the softgripper developed by Hirose & Umetani (1978), which shows great results in matching irregular shapes. Each phalanx is identical in length and shape to provide a modular option for longer or shorter fingers.

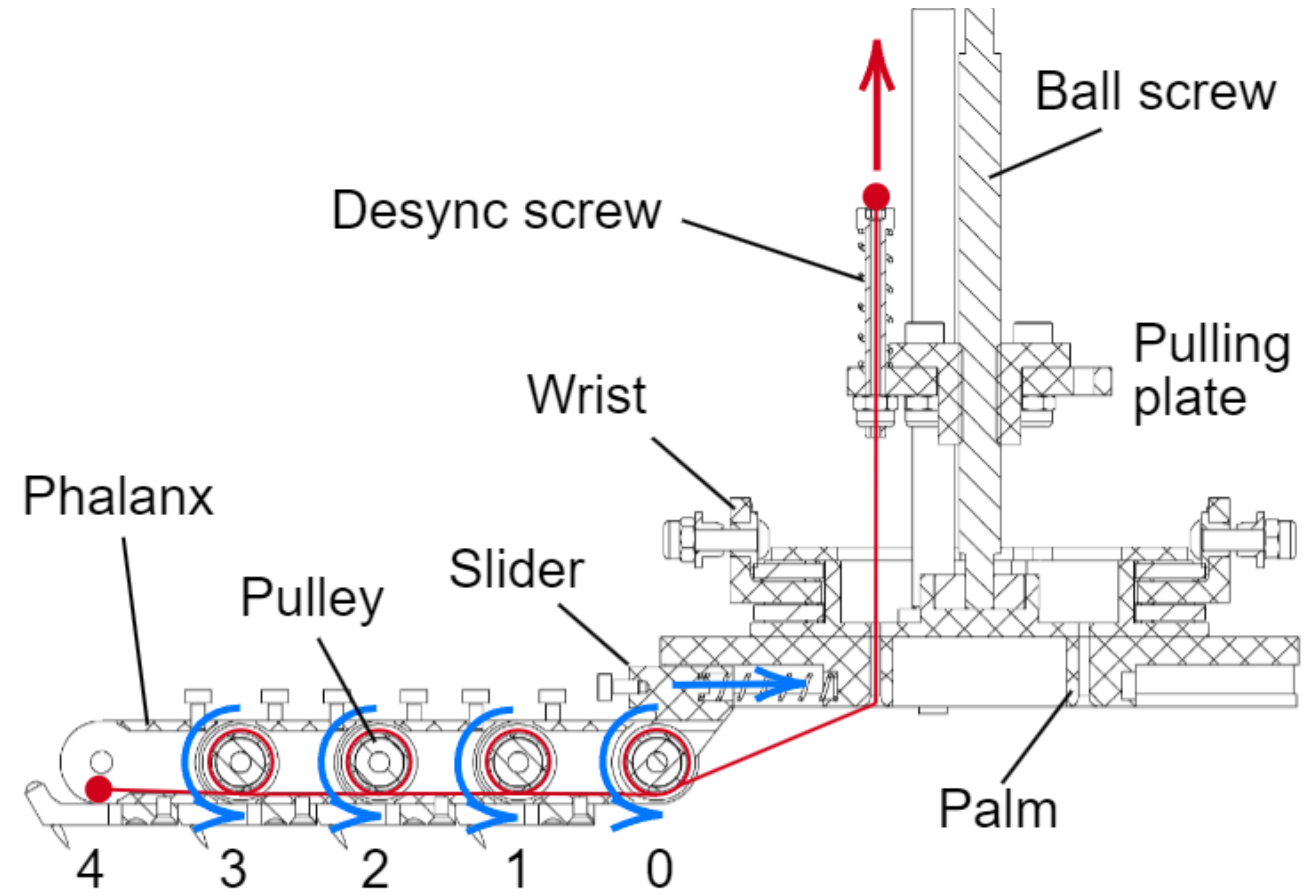

*Fig. 4 Section view of the gripping system*

The gripper is actuated by a DC brushless motor linked to a fine pitched ball screw, creating a linear motion to pull the tethers and providing a self-locking feature to reduce power consumption during grasping. The gripper opening is achieved passively by relaxing the tethers thanks to tension springs located between the phalanges. Each tether is attached to the pulling plate through a spring-loaded screw that allows compliance for the finger to individually match the local curvature of the target.

The gripper phalanges are equipped with multiple micro-spines to hook onto the asperities of a rocky surface and generate gripping. The fingers are also attached to the gripper using a prismatic joint to create a tangential motion over the surface of the target. These features increase the probability of the micro-spine to hook a suitable asperity while also creating a clamping action, therefore increasing the gripping performances.

## 2. Method

In this section, the experimental and simulation method to determine the sinkage behavior of the gripper in soft soil will be described.

The challenge of replicating an extraterrestrial environment on Earth commands us to rely on simulation models to determine the terramechanic behavior of regolith in reduced gravity. Therefore, models of the gripper and granular material were created using the Altair HyperWorks and EDEM suite, a coupling of multi-body dynamics and particle modelling. To validate the simulation model, the gripper sinkage is experimentally quantified in Toyoura sand and compared to the simulation results.

### *2.1. Gripper model*

The gripper model was created using multibody simulation software, Altair MotionSolve. Due to the highly compliant geometry of the fingers, a single solid body would not accurately demonstrate the sinkage behavior of the gripper. Therefore, a total of 41 bodies including the gripper palm, eight finger sliders, and 32 phalanges (Fig. 5) were assembled and constrained to replicate the gripper compliant properties.

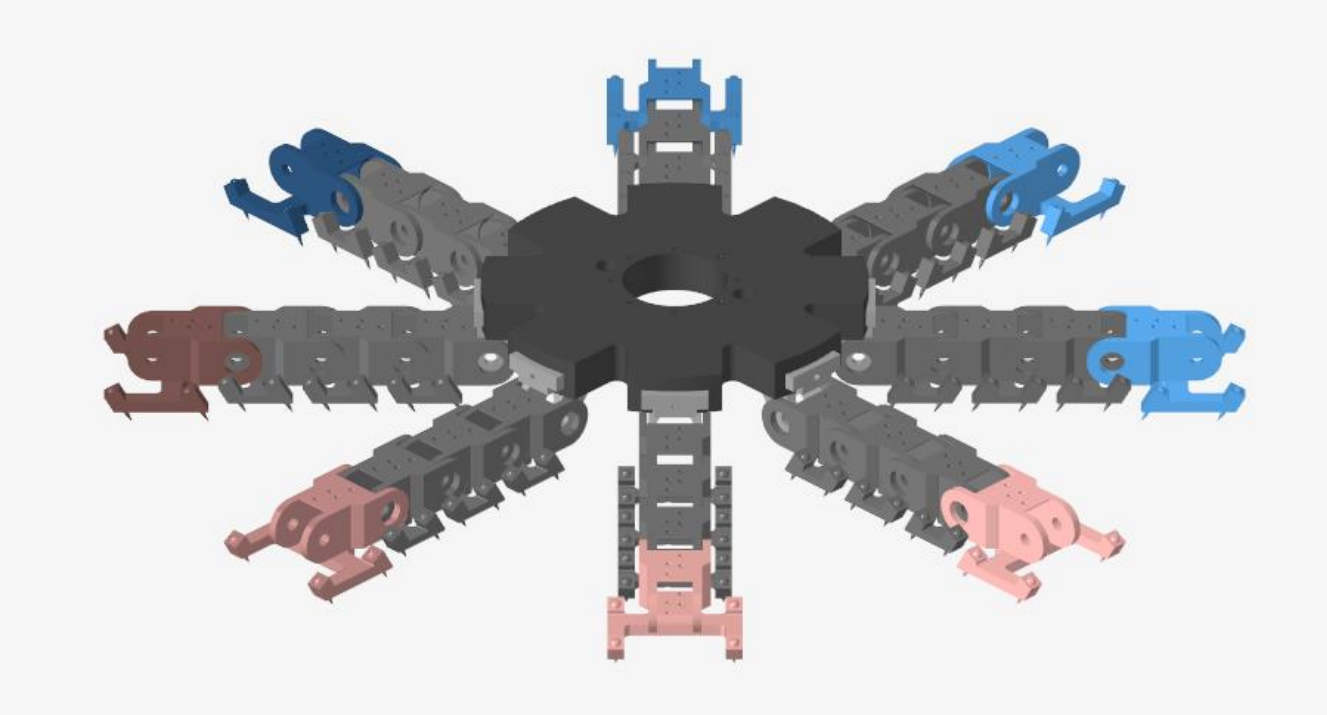

*Fig. 5 MotionSolve gripper model*

Unnecessary hardware, such as screws, bearings, axles, and clips were removed to simplify the model. The tension springs in charge of the passive opening were replaced by torsional springs centered on the phalanx joint axis. The finger slider springs were also included in the model. The tether and pulley actuation were simulated by directly applying torque onto the phalanx's joints, and motion angle limits were added to accurately replicate the compliance range of the fingers.

To recreate the dynamic motion of a step, gravity is applied to the model, as well as a load on the gripper palm that represents the robot weight using the function $\vec{\boldsymbol{L}}$ detailed in Eq. 1 and Eq. 2.

$$\vec{\boldsymbol{L}} = \begin{Bmatrix} L_x \\ L_y \\ L_z \end{Bmatrix} = \begin{Bmatrix} \sin(\theta)\sigma(t) \\ 0 \\ \cos(\theta)\sigma(t) \end{Bmatrix} \quad (1)$$

With the function *σ(t)* defined as:

$$\sigma(t) = STEP(t, t_1, mg, t_2, -5) + STEP(t, t_3, 0, t_4, -61) \quad (2)$$

The angle θ represents the slope inclination of the granular material, *t* the simulation time value, and STEP the third derivation interpolation function provided by MotionSolve.

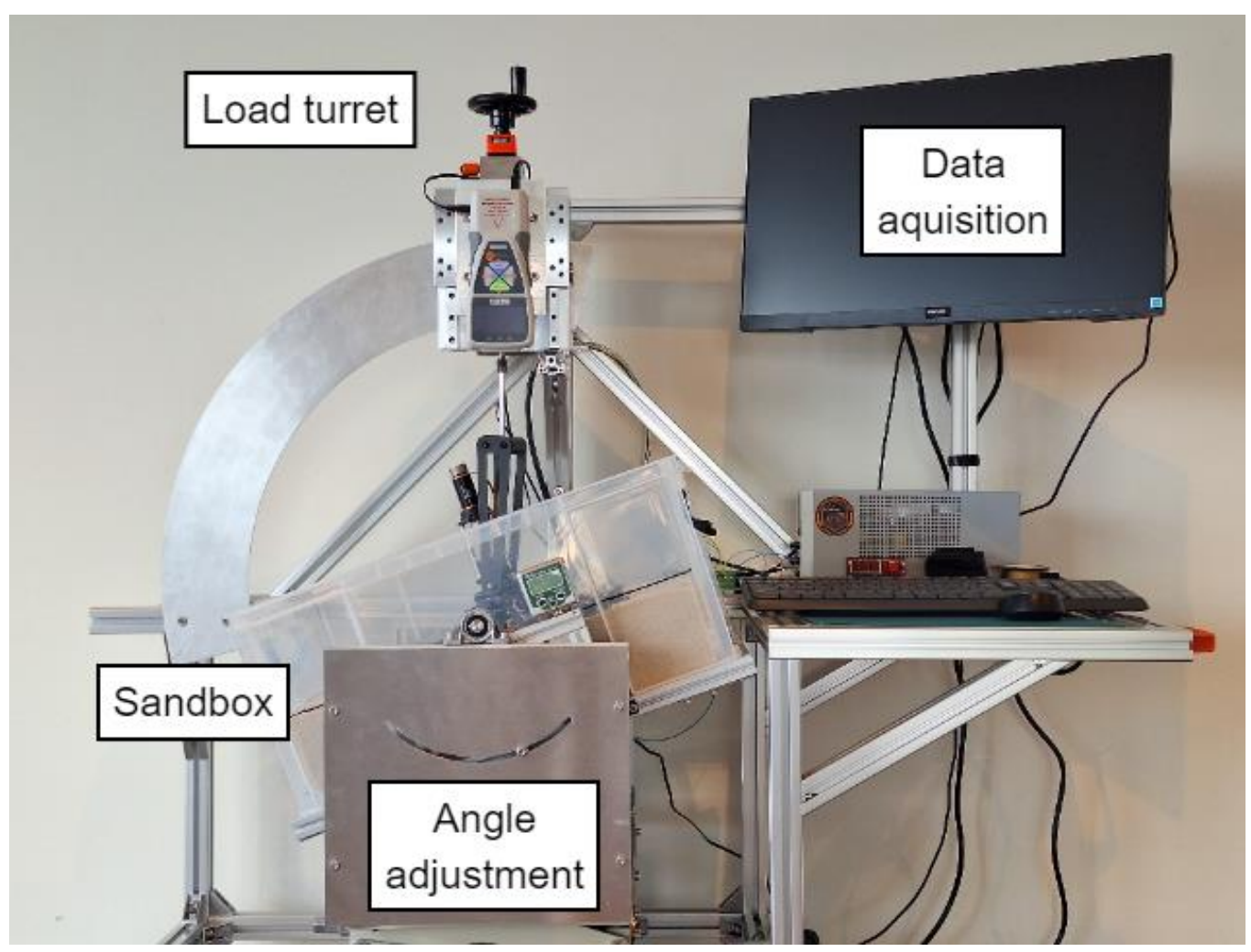

*Fig. 6 Gripper testing bench*

To avoid unwanted dynamic behavior due to the fast approach of the gripper in the granular material, the gripper weight *mg* is compensated for by the load from t0 to t1 to allow the fingers to settle in their resting position. Then the gripper contacts with the medium and stabilizes under a load of 5N from t2 to t3. Finally, the load gradually increases from t3 to t4 to the maximum of 66N, corresponding to the weight that a robot of 20kg under Earth gravity would apply to the gripper while standing on 3 legs.

### *2.2. Particle model*

The interactions between the gripper model and the particles are computed using the Discrete Element Method (DEM) provided by Altair EDEM software. The particle model was previously developed to simulate the single-wheel performance of a micro-rover, with soil parameters independently calibrated based on an angle of repose experiment and validated against single-wheel experiments before being used to study lunar behavior. The parameters for Toyoura sand, which closely mirror the silica sand used in the experiment, were validated in the prior study and are now integral to this model.

Interaction between particles or a collection of particles is simulated as soft spheres, with interactions between them defined by contact models and associated particle parameters. Two key parameters are rolling friction and static friction, accounting for the non-sphericity of actual particles. The standard Hertz-Mindlin contact model and a Type C rolling model were used (Ai et al., 2011; Coetzee and Scheffler, 2023).

*Table 1 Particle model properties for Toyoura sand and Regolith simulant*

| Parameter | Unit | Toyoura sand | Regolith |
|---|---|---|---|
| Particle size | mm | 1 | 1 |
| Particle number | | 1249000 | 1249000 |
| Density | kg/m³ | 2650 | 2857 |
| Poisson's ratio | | 0.25 | 0.3 |
| Young's Modulus | N/m² | $5e^7$ | 1e8 |
| Coefficient of Restitution | | 0.3 | 0.4 |
| Static friction | | 0.65 | 0.81 |
| Rolling friction | | 0.15 | 0.42 |
| Cohesion | N/m² | 0 | 0 |
| Angle of repose | ° | 34 | 39 |

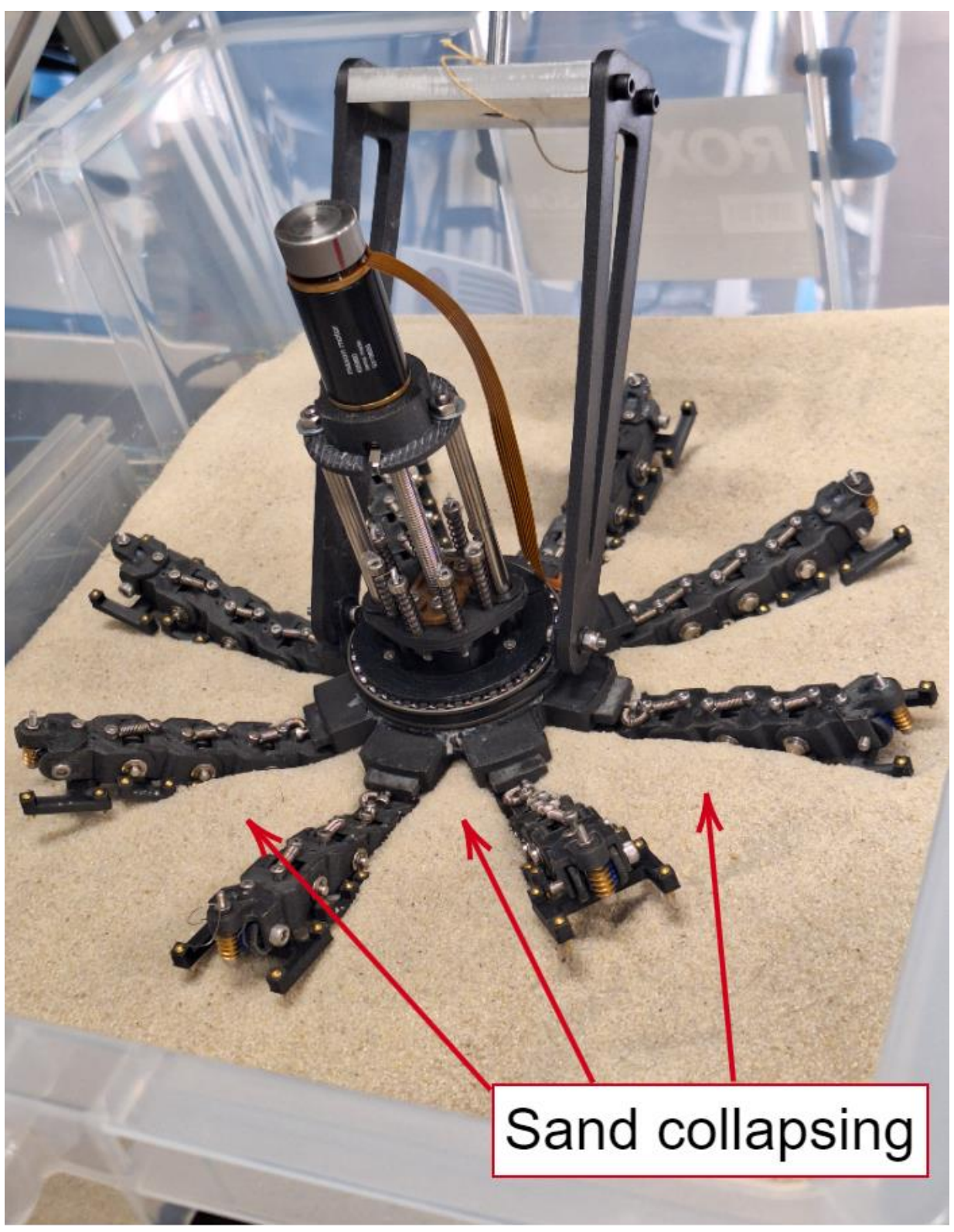

*Fig. 7 Gripper sinkage at a 35 ° slope angle*

## 2.3. Testing environment

The sinkage behavior of the gripper can be characterized by the vertical displacement required for the granular material to generate a reaction force equal to the applied load. The vertical displacement may vary depending on the properties of the granular material, the load and the angle of slope, as the material will slip and collapse. For assessing the gripper sinkage under a wide range of parameters, a testing bench, presented in Fig. 6 was developed.

The test bench is composed of a manually actuated linear turret on which is attached a force meter with a resolution of 0.01N and an acquisition frequency of 0.1s. The gripping system is connected to the force meter using a rigid assembly. The turret is mounted on a curved rail to adjust the load angle from a range of 0° to 90° that can be dialed with a precision of 0.1° using an absolute digital inclinometer. The axial displacement of the turret is measured through a graduated analog display with a resolution of 0.01mm. A sandbox is mounted on the bench using an inclination-adjustable system to simulate a ground inclination from a range of -45° to 45°. The wrist of the gripper is a universal joint with roll and yaw freedom, allowing the load to be applied at the center of the gripper regardless of inclination or orientation.

The sandbox is a 446 x 332 x 218mm container filled with silica sand to a height of 100mm from the bottom. Silica sand has been chosen for its low cohesion and the uniformity of the size of its particles. It is closely related to Toyoura sand in terms of mechanical properties and is widely available.

# 3. Results

In this section, a comparison of the experimental and simulated results of the gripper sinkage behavior will be presented.

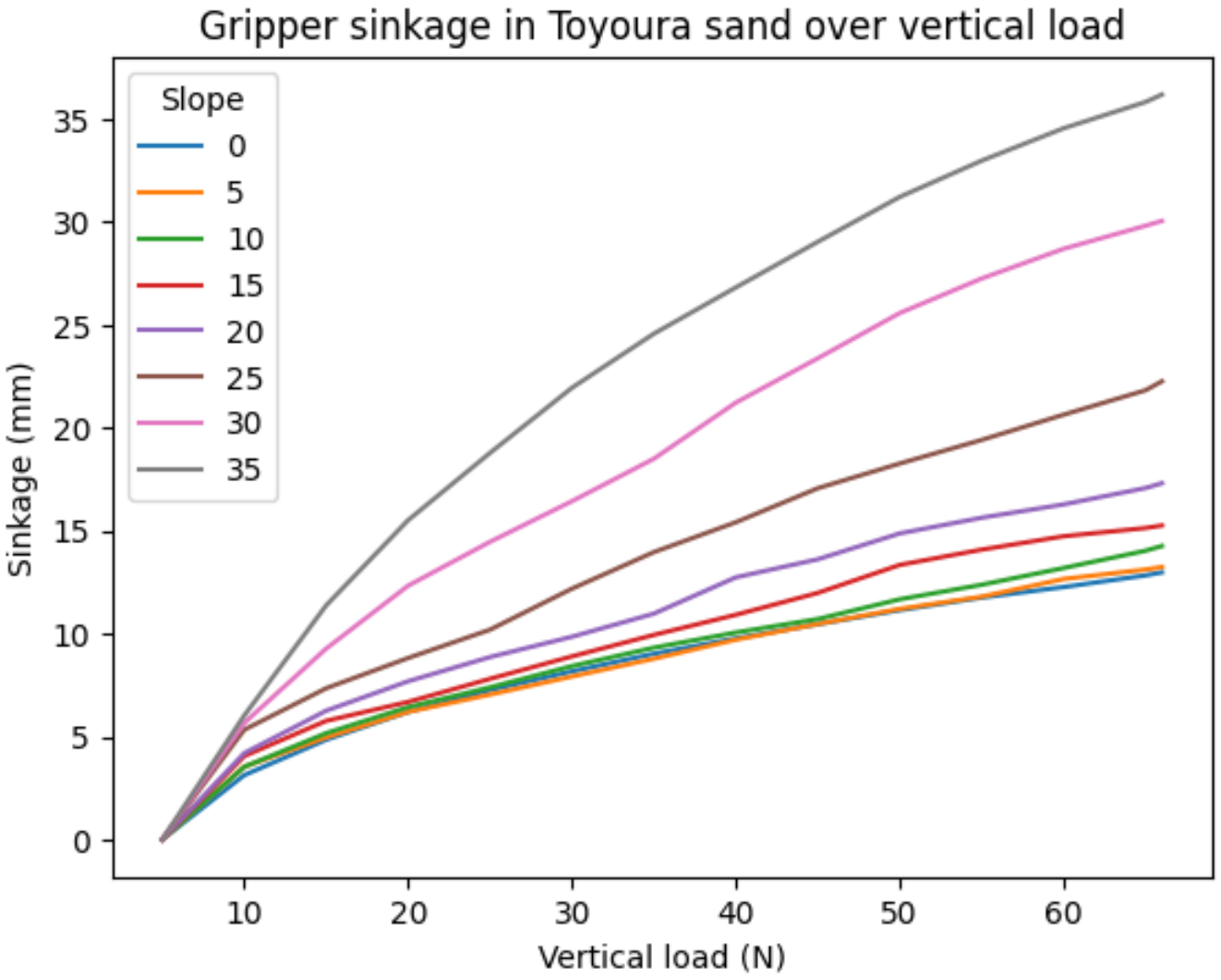

*Fig. 8 Gripper sinkage at different ground inclinations*

## 3.1. Experimental results

The sinkage of the gripper was demonstrated by first applying an initial load of 5N and recording the corresponding position of the gripper palm using the linear turret indicator. The load was then gradually increased, and the relative displacement was recorded in increments of 5N until a maximum load of 66N. To accurately represent the resting state of the gripper during walking, no tension is applied to the tethers and the fingers are free to take a more advantageous shape.

To observe the impact of the soil inclination on the sinkage, the sandbox angle was adjusted from 0° to 35° at 5° increments. Experiments above an inclination of 35° were classified as unsatisfactory as the slope angle exceeded the angle of repose of the silica sand, inducing high slipping and collapsing of the surface.

As presented in Fig. 8, the results of the experiment show that the sinkage increases while a heavier load is applied to the gripper. At a maximum of 66N, the gripper sank into the sand at an average of 12.98mm on flat ground, to 36.19mm with the ground inclined at 35°. The sinkage depth is comparable for a slope angle from 0° to 10°, but it increases greatly as the angle increases beyond 20°. At a steeper angle, the pressure generated by the gripper is not perpendicular to the ground surface, making the granular material prone to shearing and collapsing. Especially at inclinations of 30° to 35°, the granular material is close to its angle of repose and amounts of material are sheared away from the surface as the gripper digs into the surface (Fig. 7). The maximum sinkage values are presented in Table 2 and compared in percentage with the gripper diameter $D$ = 250mm.

*Table 2 Gripper maximal sinkage*

| Slope angle (°) | Sinkage (mm) | Percentage |
|---|---|---|
| 0 | 12.98 | 5.2 |
| 5 | 13.23 | 5.3 |
| 10 | 14.26 | 5.7 |
| 15 | 15.23 | 6.1 |
| 20 | 17.31 | 6.9 |
| 25 | 22.26 | 8.9 |
| 30 | 30.05 | 12 |
| 35 | 36.19 | 14.5 |

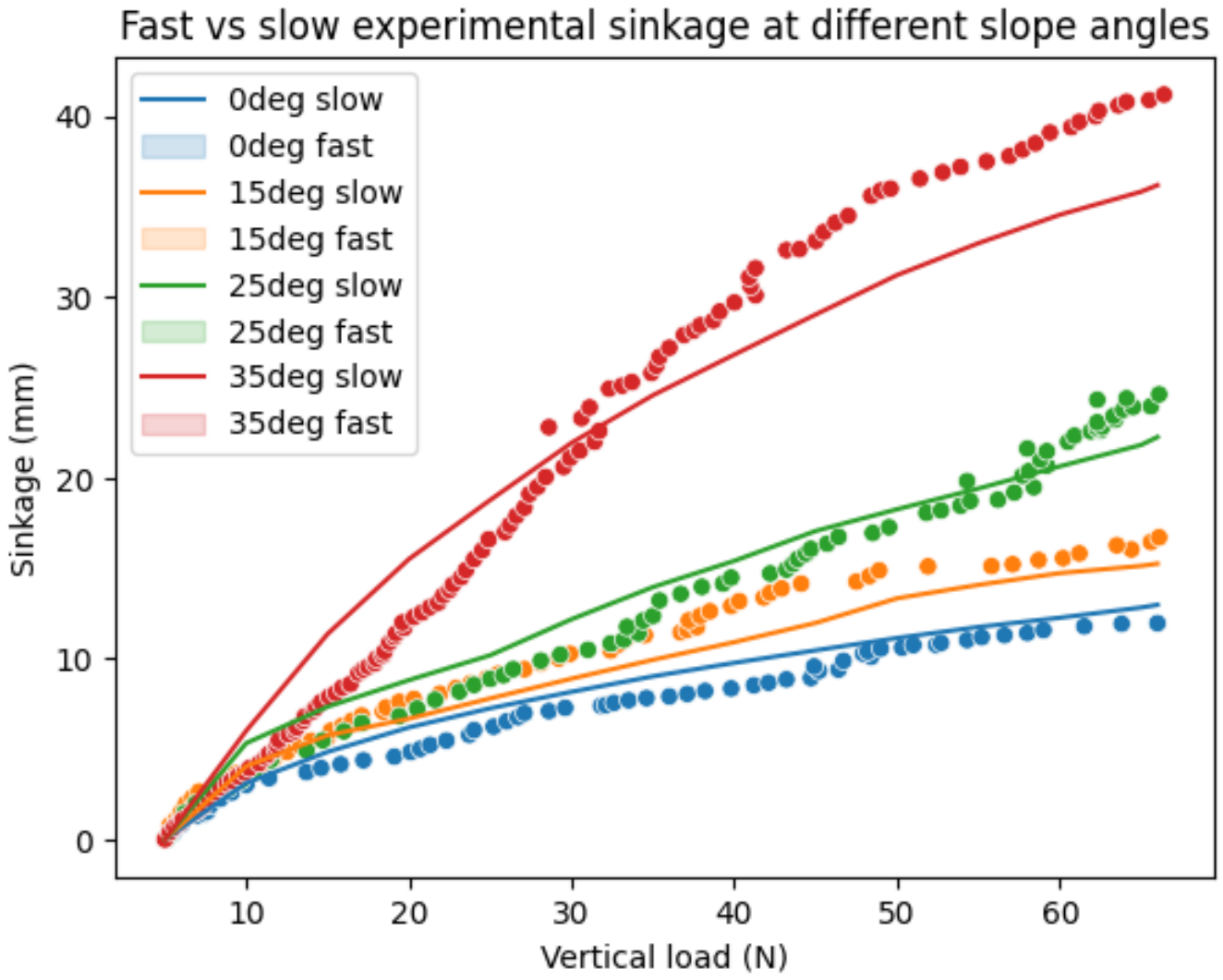

*Fig. 9 Impact of entry speed on the gripper sinkage*

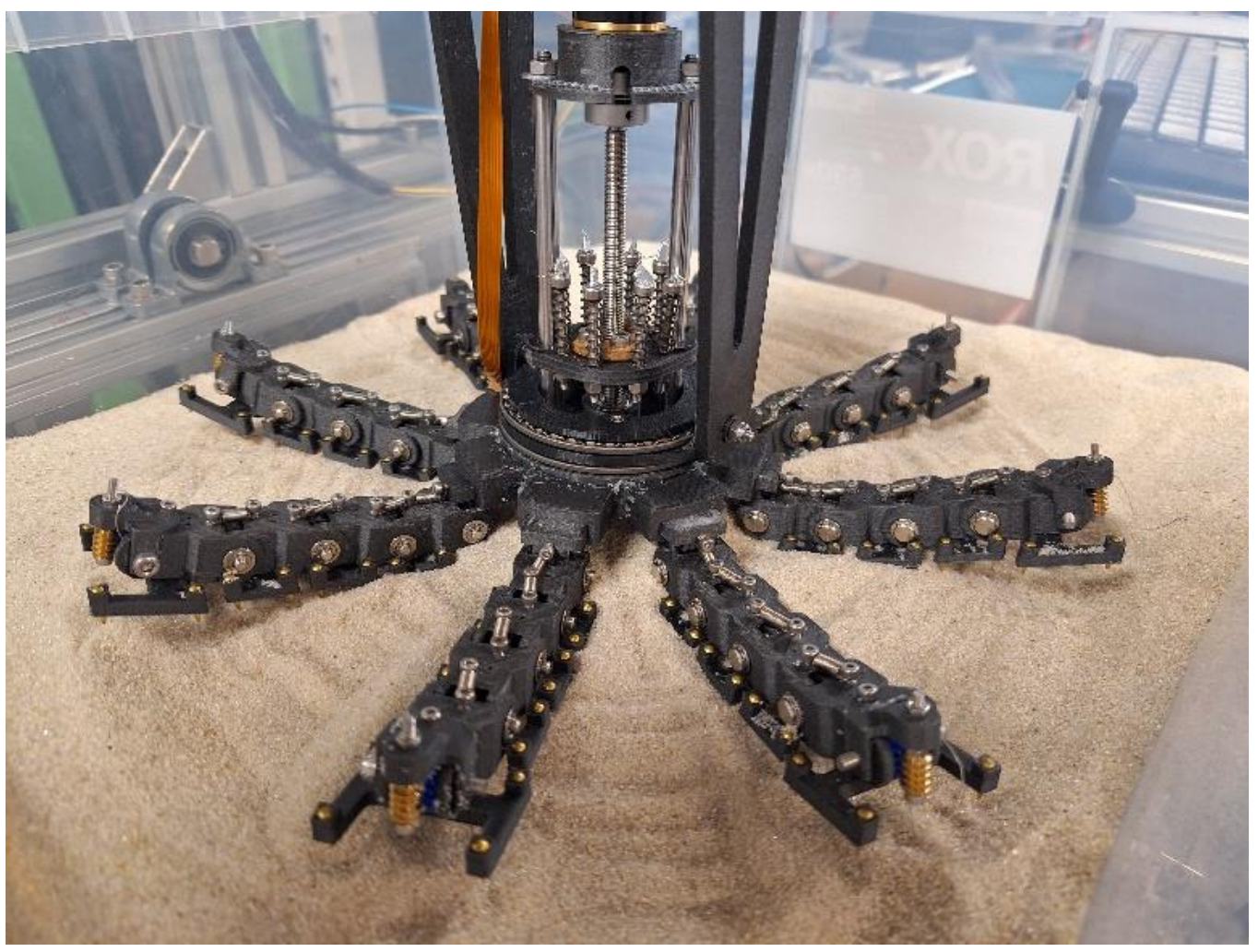

*Fig. 9 Gripper sinkage under 66N at 0°*

In the presented experiment, the manual increase of the load and recording of the displacement allows the granular material to settle and compact. On average, a single test takes approximately one minute to perform. Therefore, a second set of experiments was conducted to observe the impact of the entry speed of the gripper on the granular material. The load is aimed to continuously increase to 66N in around 10s. This represents 6 times increase in entry speed, from a range of 0.22mm/s to 0.6mm/s to a range of 1.2mm/s to 3.6mm/s.

The load was recorded using the force meter feed at a frequency of 0.1s and the vertical displacement was taken using a video camera. The sinkage comparison is presented in for a slope angle of 0°, 15°, 25°, and 35°.

The results in Fig. 9 show that the speed of entry has a moderate impact on the sinkage of the gripper. Although a faster entry does not seem to generate a deeper or shallower sinkage, an average difference of 20% can be expected. It is noted that the final depth of the sink at a load of 66N is consistent by around 10% between fast and slow entry. The highest difference is observed at a higher slope angle, especially at 35°. During experimentation, it has been remarked that a higher entry speed generated more granular material collapse, therefore increasing the sinkage depth.

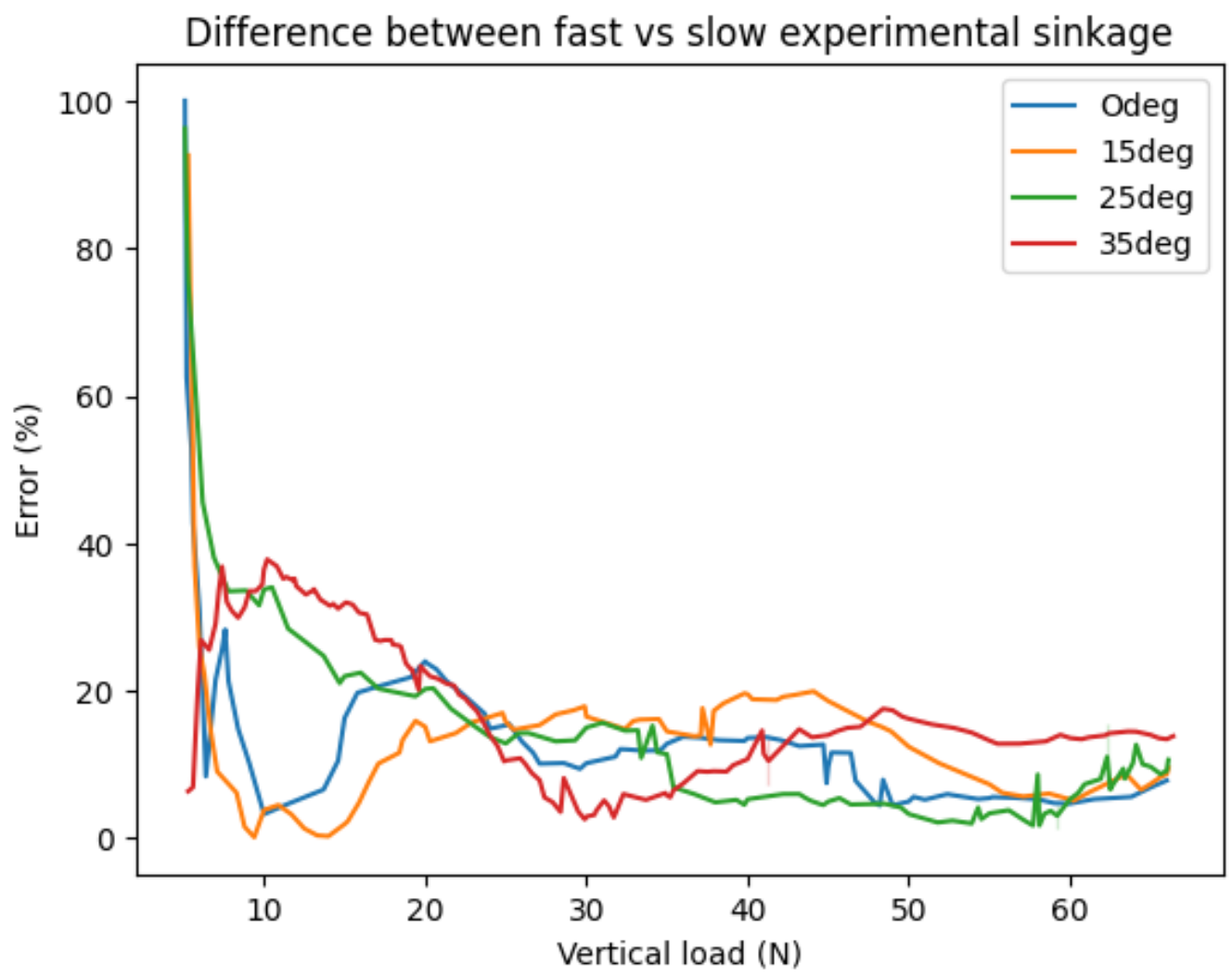

*Fig. 10 Error percentage between slow and fast entry*

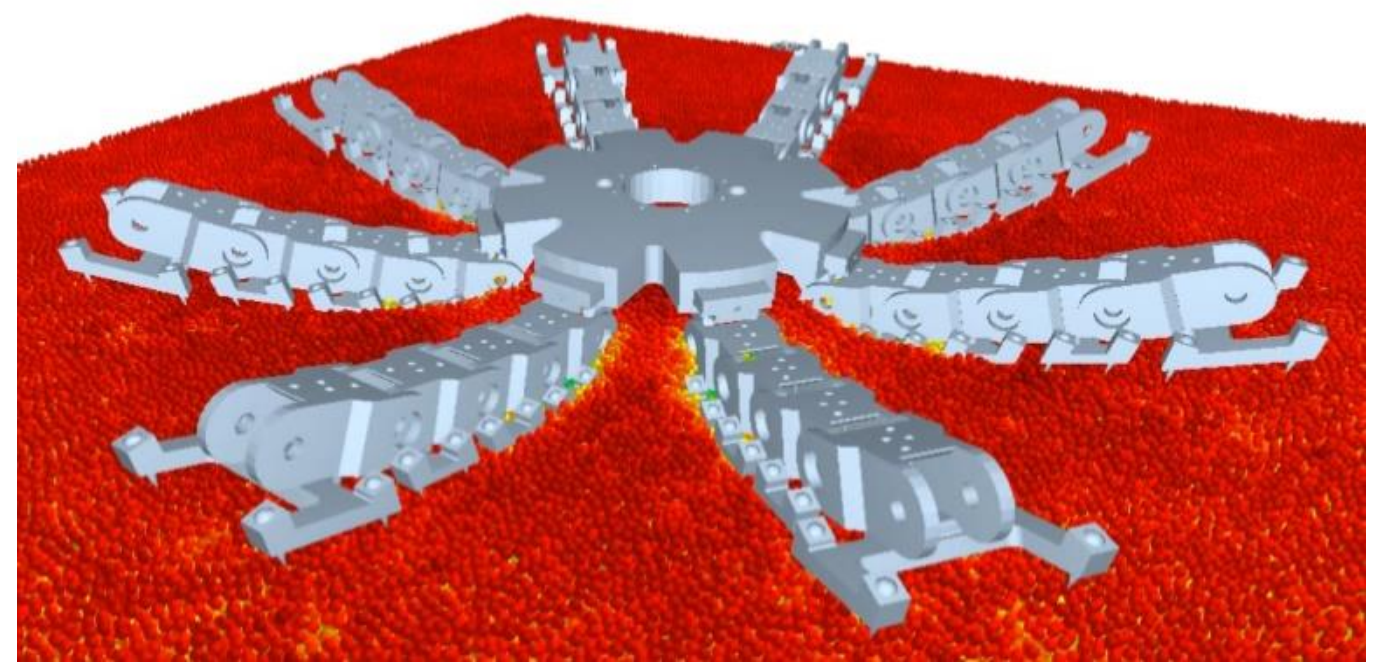

*Fig. 1 Gripper sinkage calculated in EDEM.*

### *3.2. Simulation comparison*

The gripper pressure-sinkage simulation was performed using Altair MotionSolve and EDEM software. The simulation parameters introduced in Eq. 2 were set to compute the sinkage over a period of 10s, in a manner similar to faster experimentation. The simulation required 9h of computation with a timestep of 0.01s, and the simulated sinkage can be observed against the experimental results in Fig. 13. The simulated sinkage is consistent with the experimental data, with a difference percentage mean of 13.49 % and a standard deviation of 9.15%. The most drastic difference can be observed starting at a load of 40N. Before 40N, the difference percentage mean averages 9.25% with a standard deviation of 5.61%.

*Table 3 Simulation parameters*

| Parameter | Symbol | Value |
|---|---|---|
| Simulation duration | $t_{max}$ | 10s |
| Time step | $f$ | 0.01s |
| Settling start | $t_1$ | 0 |
| Settling end | $t_2$ | 0.5 |
| Load start | $t_3$ | 0.5 |
| Load end | $t_4$ | 10 |
| Slope angle | $\theta$ | 0°, 15°, 25°, 35° |

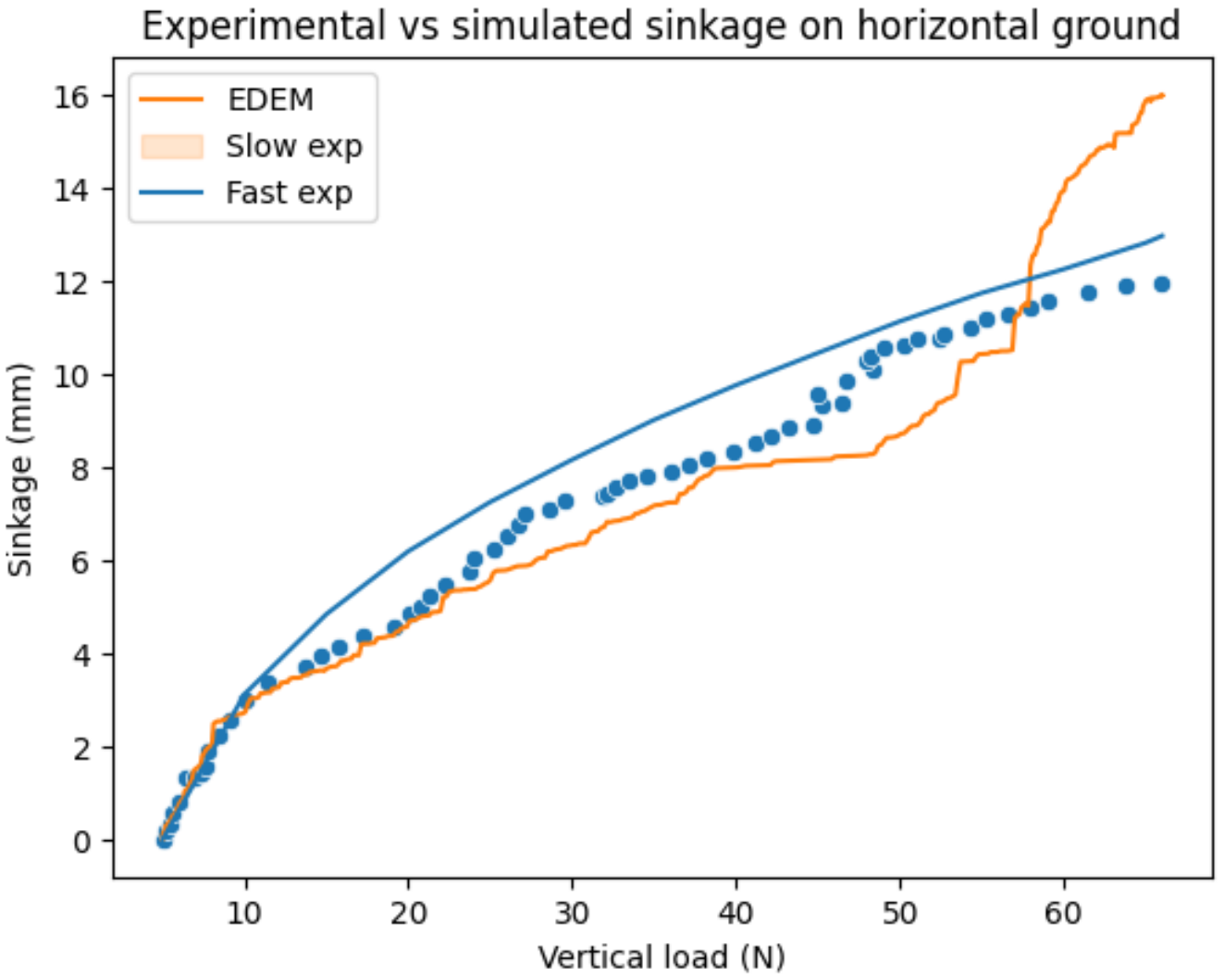

*Fig. 13 Experimental vs simulated sinkage on horizontal ground*

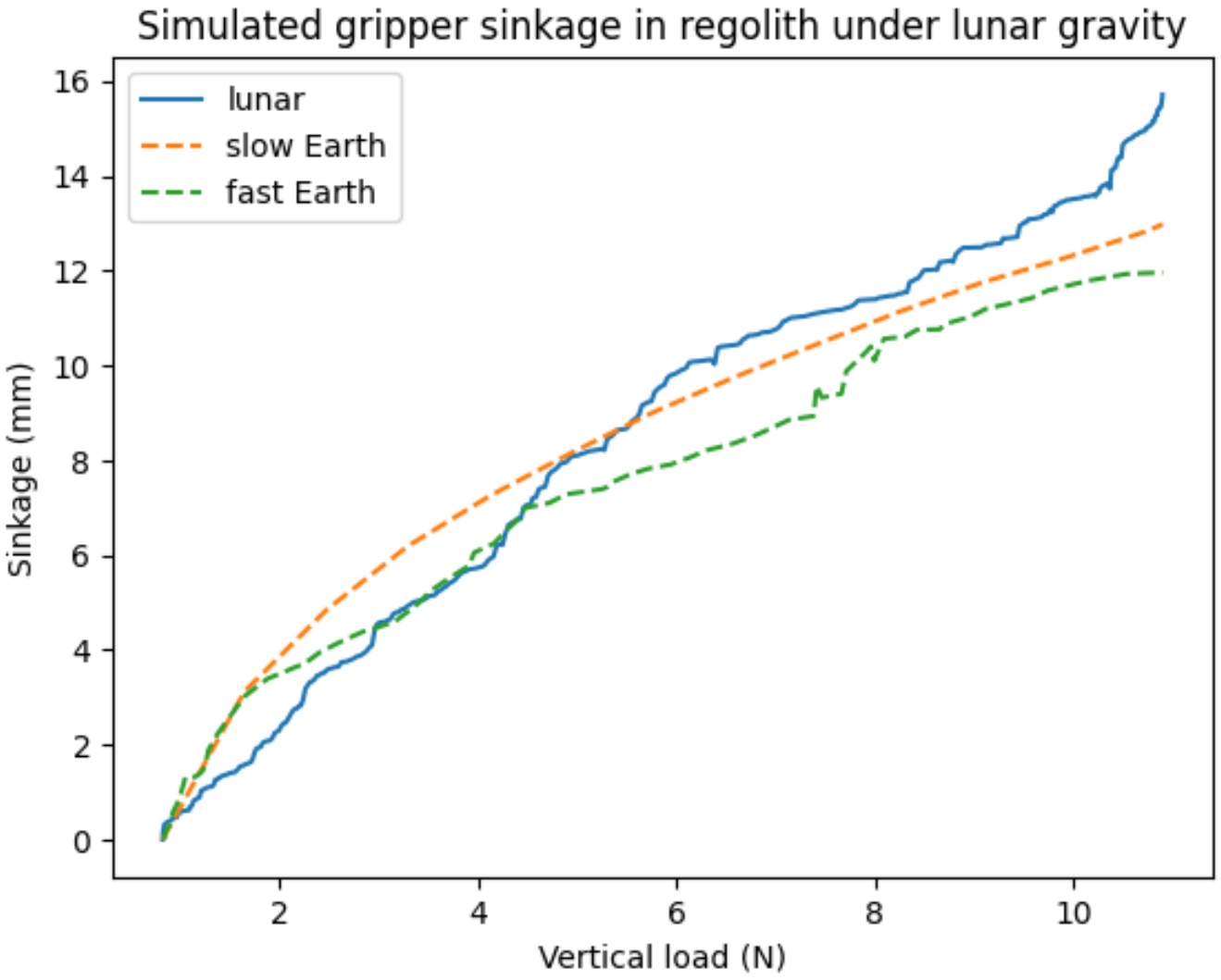

*Fig. 14 Gripper sinkage simulated in lunar environment.*

A visual appreciation of the simulated results reveals that some particles clip through the gripper model at higher load. The sharp increases in sinkage after 50N are evidence that particle clipping occurs and voids in the granular material are quickly filled by other particles, resulting in deeper sinkage. An increase in the particle density of the simulated sandbox is considered, although this would exponentially lengthen the computation time. The gripper model may also be simplified further by removing unnecessary geometries.

### *3.3. Space environment simulation*

Figure 14 introduces the sinkage computed in lunar gravity using the regolith particle model introduced in 2.2. As a comparison, the data collected from the experiments on Earth were correlated to lunar gravity by adjusting the force values by the factor $s = \frac{1.62}{9.81}$. It can be observed that the simulated sinkage values are similar to the experimental values, as shown by the difference estimation presented in Table 4.

*Table 4 Difference between lunar simulation and experimental results*

| Experiment | Mean (%) | SD (%) | Min (mm) | Max (mm) |
|---|---|---|---|---|
| Slow entry | 13.71 | 12.53 | 0.01073 | 2.7396 |
| Fast entry | 13.35 | 7.32 | 0.0019 | 3.7531 |

Although the granular material properties are different between Toyoura sand and regolith, it is possible to consider that gravity does not hold a high impact on the sinkage behavior of the gripper. In lower gravity environments, the robot weight applied to the gripper is greatly reduced, which would result in a lower sinkage than on Earth. However, gravity also impacts the granular material particle interactions, causing the soil to be less dense and compact, therefore inducing higher sinkage. These two factors are believed to compensate for each other, explaining the similarity in sinkage between terrestrial and lunar environments.

Kobayashi et al. (2010)present similar conjectures regarding the effect of gravity on wheel sinkage, although reduced gravity appears to have a negative impact on traction performance, reinforcing the argument that alternative means of locomotion need to be developed.

Nonetheless, it can be concluded that sinkage experiments in Toyoura sand or regolith under Earth gravity can offer a rough estimate of the expected sinkage in lunar environment.

## 4. Discussion

This study demonstrates that a legged robot end effector is subject to significant sinkage while walking in a granular environment. The sinkage is directly affected by the properties of the granular material, the load, the inclination of the ground, and gravity.

The velocity of entry in the granular material moderately impacts the sinkage behavior of the gripper and must be taken into consideration for qualitative estimations, particularly on steeper ground. For the case of wheeled vehicles and robots, multiple pressure-sinkage models such as Bekker and Reece propose a logarithmic correlation between pressure and sinkage based on material properties. Then it can be discussed that to reduce the sinkage, the gripper palm could be modified to increase the contact surface, thus reducing the pressure applied to the granular material. In addition, flexible webbing between the fingers could also be integrated if the grasping performances are not hindered.

Although the experiments yield satisfactory and repeatable results, improvements can be made by integrating an automated linear actuator to control the sinkage rate and displacement of the gripper. Such a device can be directly coupled with the force meter to generate more precise results, but also reproduce more complex situations such as exponentially increasing load, variable velocity entry, stomping and so on.

The gripper sinkage simulation holds promising results, and accuracy of the model can be confirmed by comparison with the experimental data. The issue of particle clipping needs to be addressed to completely validate the simulation model, and comparisons with different slope angles are to be conducted.

Finally, the method can also be tested using granular materials more appropriate to observe the sinkage of the gripper in space environment, such as regolith simulant. Based on the simulation results in lunar conditions, the sinkage is seemingly close to the experimental values on Earth gravity, and further experiments are required to confirm the similarity. Tests can also be conducted in Martian conditions to observe the extent of this correlation.

Regolith simulant is extremely thin compared to silica sand and requires measures to safely conduct experiments and protect the gripper from infiltration and clogging. An origami-based protective cover made of polyvinyl fluoride is currently being developed to seal the gripper from its environment while still offering full mobility and compliance of the fingers.

## 5. Conclusion

This research introduces the terramechanic behavior of an adaptive gripping system for legged robots to navigate and operate on extraterrestrial surfaces, addressing the limitations of wheeled rovers in accessing deeper lunar areas such as caves and craters.

Experimental observations of the gripper's sinkage in Toyoura sand, coupled with simulations using Altair MotionSolve and EDEM, provided valuable insights into to the gripper terramechanic performance on granular terrain. The simulation model extended the analysis to lunar gravity conditions, enhancing relevance to actual lunar missions.

The findings indicate that the gripper's sinkage can be predicted and integrated into the motion control algorithms of legged robots, potentially improving their mobility and operational accuracy on granular surfaces. This study lays the groundwork for enhancing the functionality of robotic systems in extraterrestrial environments, with future research aimed at optimizing the gripper design and refining the simulation models for broader lunar conditions.

## 6. Acknowledgement

This work was supported by the Japan Science and Technology Agency "Support for Pioneering Research Initiated by the Next Generation", Asteroid Mining Corporation Ltd., GP-Mech Program of Tohoku University, Japan and the Space Robotics Laboratory of Tohoku University, Japan.

## 7. Declaration of competing interest

The authors declare that they have no known competing financial interests or personal relationships that could have appeared to influence the work reported in this paper.

## 8. References

Ai, J., Chen, J.-F., Rotter, J.M., Ooi, J.Y., 2011. Assessment of rolling resistance models in discrete element simulations. Powder Technol 206, 269–282. https://doi.org/https://doi.org/10.1016/j.powtec.2010.09.030

Bekker, M.G., Kitano, M., Kuma, A., 1962. Theory of Land Locomotion: The Mechanics of Vehicle Mobility. University of Michigan Press.

Candalot, A., Hashim, MM., Hickey, B., Laine, M., Hunter-Scullion, M., Yoshida, K. (2024). Soft Gripping System for Space Exploration Legged Robots. In: Berns, K., Tokhi, M.O., Roennau, A., Silva, M.F., Dillmann, R. (eds) Walking Robots into Real World. CLAWAR 2024. Lecture Notes in Networks and Systems, vol 1115. Springer, Cham. https://doi.org/10.1007/978-3-031-71301-9_14

Coetzee, C., Scheffler, O.C., 2023. Comparing particle shape representations and contact models for DEM simulation of bulk cohesive behaviour. Comput Geotech 159, 105449. https://doi.org/https://doi.org/10.1016/j.compgeo.2023.105449

Di Carlo, J., Wensing, P.M., Katz, B., Bledt, G., Kim, S., 2018. Dynamic Locomotion in the MIT Cheetah 3 Through Convex Model-Predictive Control, in: 2018 IEEE/RSJ International Conference on Intelligent Robots and Systems (IROS). pp. 1–9. https://doi.org/10.1109/IROS.2018.8594448

Graf Nicole M. and Behr, A.M. and D.K.A., 2019. Crab-Like Hexapod Feet for Amphibious Walking in Sand and Waves, in: Martinez-Hernandez Uriel and Vouloutsi, V. and M.A. and M.M. and A.M. and P.T.J. and V.P.F.M.J. (Ed.), Biomimetic and Biohybrid Systems. Springer International Publishing, Cham, pp. 158–170.

Hirose, S., Umetani, Y., 1978. The development of soft gripper for the versatile robot hand. Mech Mach Theory 13, 351–359. https://doi.org/https://doi.org/10.1016/0094-114X(78)90059-9

Ishigami, G., Miwa, A., Nagatani, K., Yoshida, K., 2006. Terramechanics-based analysis on slope traversability for a planetary exploration rover. Proceedings of the Twenty-Fifth International Symposium on Space Technology and Science, Kanazawa 2006.

Kobayashi, T., Fujiwara, Y., Yamakawa, J., Yasufuku, N., Omine, K., 2010. Mobility performance of a rigid wheel in low gravity environments. J Terramech 47, 261–274. https://doi.org/https://doi.org/10.1016/j.jterra.2009.12.001

Komizunai, S., Konno, A., Abiko, S., Uchiyama, M., 2010. Development of a static sinkage model for a biped robot on loose soil, in: 2010 IEEE/SICE International Symposium on System Integration. pp. 61–66. https://doi.org/10.1109/SII.2010.5708302

Li, S., Milliken, R.E., 2017. Water on the surface of the Moon as seen by the Moon Mineralogy Mapper: Distribution, abundance, and origins. Sci Adv 3, e1701471. https://doi.org/10.1126/sciadv.1701471

Marvi, H., Gong, C., Gravish, N., Astley, H., Travers, M., Hatton, R.L., Mendelson, J.R., Choset, H., Hu, D.L., Goldman, D.I., 2014. Sidewinding with minimal slip: Snake and robot ascent of sandy slopes. Science (1979) 346, 224–229. https://doi.org/10.1126/science.1255718

Nagaoka, K., Sawada, K., Yoshida, K., 2020. Shape effects of wheel grousers on traction performance on sandy terrain. J Terramech 90, 23–30. https://doi.org/https://doi.org/10.1016/j.jterra.2019.08.001

Oettershagen, P., Lew, T., Tardy, A., Michaud, S., 2019. Investigation of specific wheel-terrain interaction aspects using an advanced single wheel test facility. ASTRA 2019.

Parness, A., Abcouwer, N., Fuller, C., Wiltsie, N., Nash, J., Kennedy, B., 2017. LEMUR 3: A limbed climbing robot for extreme terrain mobility in space, in: 2017 IEEE International Conference on Robotics and Automation (ICRA). pp. 5467–5473. https://doi.org/10.1109/ICRA.2017.7989643

Rodríguez-Martínez, D., Uno, K., Sawa, K., Uda, M., Kudo, G., Diaz, G.H., Umemura, A., Santra, S., Yoshida, K., 2024. Enabling Faster Locomotion of Planetary Rovers With a Mechanically-Hybrid Suspension. IEEE Robot Autom Lett 9, 619–626. https://doi.org/10.1109/LRA.2023.3335769

Saranli, U., Buehler, M., Koditschek, D., 2001. RHex: A Simple and Highly Mobile Hexapod Robot. Departmental Papers (ESE) 20.

Uno, K., Takada, N., Okawara, T., Haji, K., Candalot, A., Ribeiro, W.F.R., Nagaoka, K., Yoshida, K., 2021. HubRobo: A Lightweight Multi-Limbed Climbing Robot for Exploration in Challenging Terrain, in: 2020 IEEE-RAS 20th International Conference on Humanoid Robots (Humanoids). pp. 209–215. https://doi.org/10.1109/HUMANOIDS47582.2021.9555799

Welch, R., Limonadi, D., Manning, R., 2013. Systems engineering the Curiosity Rover: A retrospective, in: 2013 8th International Conference on System of Systems Engineering. pp. 70–75. https://doi.org/10.1109/SYSoSE.2013.6575245

Wong, J.-Y., Reece, A.R., 1967. Prediction of rigid wheel performance based on the analysis of soil-wheel stresses part I. Performance of driven rigid wheels. J Terramech 4, 81–98. https://doi.org/https://doi.org/10.1016/0022-4898(67)90105-X

Yang, H., Zhang, C., Ding, L., Wei, Q., Gao, H., Liu, G., Ge, L., Deng, Z., 2024. Comparative study of terramechanics properties of spherical and cylindrical feet for planetary legged robots on deformable terrain. J Terramech 113–114, 100968. https://doi.org/https://doi.org/10.1016/j.jterra.2024.100968